\documentclass[conference]{IEEEtran}
\IEEEoverridecommandlockouts

\usepackage{cite}
\usepackage{amsmath,amssymb,amsfonts}
\usepackage{algorithmic}
\usepackage{algorithm}
\usepackage{graphicx}
\usepackage{array}
\usepackage{textcomp}
\usepackage{url} % Typesets URLs without creating clickable PDF annotations
\usepackage[T1]{fontenc} % Underscore in bibtex url
\usepackage{color,soul}
\usepackage{subcaption}
\usepackage{dcolumn}
\usepackage{stackengine}
\usepackage{booktabs}

\usepackage{booktabs,tabulary,lipsum}
\usepackage{dcolumn,tipa}
\newcolumntype{d}{D{.}{.}{7}}
\usepackage{siunitx}

\usepackage{adjustbox, booktabs, multirow}

\def\BibTeX{{\rm B\kern-.05em{\sc i\kern-.025em b}\kern-.08em
    T\kern-.1667em\lower.7ex\hbox{E}\kern-.125emX}}

\usepackage[table,xcdraw,usenames,dvipsnames]{xcolor}

\usepackage{amsmath}
\usepackage{multirow}

\usepackage[final]{changes}
\newcommand{\stkout}[1]{\ifmmode\text{\sout{\ensuremath{#1}}}\else\sout{#1}\fi}
\setdeletedmarkup{\stkout{#1}}

\usepackage{amsthm}

\theoremstyle{definition}

\makeatletter
\newcommand{\linebreakand}{%
  \end{@IEEEauthorhalign}
  \hfill\mbox{}\par
  \mbox{}\hfill\begin{@IEEEauthorhalign}
}
\makeatother

\usepackage[edges]{forest}
\definecolor{folderbg}{RGB}{124,166,198}
\definecolor{folderborder}{RGB}{110,144,169}
\newlength\Size
\tikzset{%
  folder/.pic={%
    \filldraw [draw=folderborder, top color=folderbg!50, bottom color=folderbg] (-1.05*\Size,0.2\Size+5pt) rectangle ++(.75*\Size,-0.2\Size-5pt);
    \filldraw [draw=folderborder, top color=folderbg!50, bottom color=folderbg] (-1.15*\Size,-\Size) rectangle (1.15*\Size,\Size);
  },
  file/.pic={%
    \filldraw [draw=folderborder, top color=folderbg!5, bottom color=folderbg!10] (-\Size,.4*\Size+5pt) coordinate (a) |- (\Size,-1.2*\Size) coordinate (b) -- ++(0,1.6*\Size) coordinate (c) -- ++(-5pt,5pt) coordinate (d) -- cycle (d) |- (c) ;
  },
}
\forestset{%
  declare autowrapped toks={pic me}{},
  pic dir tree/.style={%
    for tree={%
      folder,
      font=\ttfamily,
      grow'=0,
    },
    before typesetting nodes={%
      for tree={%
        edge label+/.option={pic me},
      },
    },
  },
  pic me set/.code n args=2{%
    \forestset{%
      #1/.style={%
        inner xsep=2\Size,
        pic me={pic {#2}},
      }
    }
  },
  pic me set={directory}{folder},
  pic me set={file}{file},
}

\usepackage{listings}
\lstdefinelanguage{Solidity}{
    keywords={function, returns, if, else, return, require, emit, public, internal, address, uint256, bool, true, false},
    keywordstyle=\color{blue}\bfseries,
    ndkeywords={},
    ndkeywordstyle=\color{orange}\bfseries,
    identifierstyle=\color{black},
    sensitive=false,
    comment=[l]{//},
    morecomment=[s]{/*}{*/},
    commentstyle=\color{gray}\ttfamily,
    stringstyle=\color{red}\ttfamily,
    morestring=[b]",
    morestring=[b]',
    morecomment=[s]{/*}{*/},
    tabsize=2,
    basicstyle=\footnotesize\ttfamily,
    numbers=left,
    numberstyle=\tiny,
    stepnumber=1,
    numbersep=5pt,
    breaklines=true,
    showstringspaces=false
}

\begin{document}

\title{
% Navigating the Crypto Waters: Machine Learning for Rug Pull Detection in Solana Memecoins
% Catching the Rug: Machine Learning for Identifying Rug Pulls in Solana Memecoins
Catching the Rug: Early Prediction of Fraudulent Memecoins on Solana via Machine Learning
}

% double-blind
\author{
\IEEEauthorblockN{Jianghai Li}
\IEEEauthorblockA{\textit{Higher School of Economics}\\
Moscow, Russia 
% \\
% li.tszyankhay@hse.ru
}
\and
\IEEEauthorblockN{Pavel Kuznetsov}
\IEEEauthorblockA{\textit{Moscow State University}\\
Moscow, Russia 
% \\
% pavelkuznetcov2002@gmail.com
}
\and
\IEEEauthorblockN{Pattapon Tanankakorn}
\IEEEauthorblockA{\textit{Skolkovo Institute of Science and Technology}\\
Moscow, Russia 
% \\
% pattapon.tanankakorn@skoltech.ru
}
\linebreakand
\IEEEauthorblockN{Konstantin Nott-Whaley}
\IEEEauthorblockA{
\textit{Independent Researcher} \\
London, UK 
% \\
% codingwitch@protonmail.com
}
\and
\IEEEauthorblockN{Igor Vodolazov}
\IEEEauthorblockA{
\textit{Independent Researcher} \\
Barcelona, Spain 
% \\
% allcryptodata@proton.me
}
\and
\IEEEauthorblockN{Yury Yanovich}
\IEEEauthorblockA{\textit{Skolkovo Institute of Science and Technology}\\
Moscow, Russia 
% \\
% y.yanovich@skoltech.ru
}
% \affiliation[HSE]{organization={Faculty of Computer Science, HSE University},
%             city={Moscow},
%             country={Russia}}
}

\maketitle

\begin{abstract}
The rapid proliferation of memecoins on blockchain platforms has increased the risk of fraudulent activities, particularly rug pulls. While previous studies have focused on Ethereum-based tokens, this paper shifts the spotlight to Solana, the leading blockchain for memecoins by trading volume and token count. Unlike Ethereum, where rug pulls often exploit smart contract backdoors, Solana memecoin rug pulls are predominantly driven by liquidity manipulation and social dynamics. This research pioneers large-scale rug pull early detection in the Solana ecosystem by assembling a dataset of 6.4 million tokens over 7 months. Market analysis reveals that a vast majority of these memecoins exhibit rug pull characteristics within one hour of launch, highlighting the urgency of short-horizon prediction. Despite the absence of code-level features, we demonstrate that classic machine learning models, particularly Gradient Boosting (XGBoost), achieve robust performance in detecting potential rug pulls using only the first 5 minutes of trading data. Furthermore, we evaluate cross-platform generalization between PumpFun and Raydium, revealing that multi-source data fusion significantly mitigates domain shift and improves detection reliability. This study advances the understanding of DeFi fraud on high-throughput chains and provides a practical framework for protecting investors.
\end{abstract}


\section{Introduction}
\label{section:intro}

The rapid advancement of blockchain technology, Web3 innovations, and decentralized finance (DeFi) has significantly fueled the expansion of the cryptocurrency market \cite{Meyer2022}. By May 2024, the market capitalization of cryptocurrencies surpassed 2 trillion US dollars \cite{CoinMarketCap2024}, highlighting the profound influence of these technologies. To support token transactions and provide alternatives to peer-to-peer exchanges, various trading platforms and cryptocurrency exchanges have emerged. These platforms are categorized into centralized exchanges (CEXs) \cite{Lee2023}, which operate similarly to traditional financial institutions, and decentralized exchanges (DEXs) \cite{DosSantos2022}, which utilize smart contracts and cryptographic methods to ensure asset security.

The rise in DEX activity has significantly impacted the memecoin market. In 2024, memecoins emerged as a leading trend in the crypto world, not generating the largest trade volume overall, but driving substantial activity on DEXs \cite{CMCResearch2024}. Memecoins are a type of crypto asset often created as a joke or to capitalize on memes and trends on the Internet \cite{Stencel2023}. Unlike traditional cryptocurrencies such as Bitcoin or Ethereum, which have specific use cases and underlying technologies, memecoins typically lack inherent utility or serious purpose. Instead, they gain popularity through social networks, community engagement, and viral marketing. Modern memecoins are tokens deployed on established blockchains (e.g., Solana, Ethereum) rather than native coins with dedicated consensus layers. However, this growth has led to an increase in fraudulent schemes, such as rug pulls \cite{Chainalysis2024,Zhou2024}, where developers promote a token to attract investors and then withdraw liquidity, causing the token's value to plummet. The decentralized nature of DEXs and user anonymity further complicate the detection of such schemes. As DeFi continues to expand, developing effective methods to detect rug pulls on Solana and other blockchains is crucial to safeguard users and maintain trust in the ecosystem.

Rug pulls on decentralized exchanges pose a significant threat to investor confidence and market stability. Currently, there is a lack of specialized tools for real-time rug pull detection on DEXs, allowing fraudsters to exploit this vulnerability. Consequently, there is an urgent need for mechanisms capable of identifying potential rug pulls before they occur~\cite{Mazorra2022,Srifa2025}.

Although artificial intelligence has become a cornerstone of financial fraud detection~\cite{yang2026review}, rug pulls in decentralized memecoin markets present unique challenges: extreme class imbalance, rapid token lifecycles, and chain-specific microstructures not addressed by traditional frameworks. Recent work has begun to address these challenges on specific chains: Hu et al.~\cite{hu2026memetrans} introduced \textit{MemeTrans}, a 40k Solana memecoin dataset focused on launchpad-phase risk annotation, while Yaremus et al.~\cite{yaremus2025ton} developed a TVL/Idle-based detection framework for the TON blockchain. Our work extends this line of research by scaling to 6.4M Solana tokens over 7 months, emphasizing early prediction and evaluating cross-platform generalization between PumpFun and Raydium. The main contributions of this paper are summarized as follows:
\begin{itemize}
    \item We assemble and analyze a comprehensive dataset of 6.4 million Solana memecoins over a 7-month period, exceeding prior Solana-specific studies by over 160$\times$ in scale, which enables robust statistical analysis of rare rug pull patterns. We are also open-sourcing some datasets (https://www.kaggle.com/datasets/btclee/memecoins).
    \item We conduct a rigorous evaluation of model transferability between Solana's primary memecoin platforms, PumpFun and Raydium, revealing significant distribution shifts and demonstrating that multi-source data fusion combined with tree-based models yields the most robust cross-platform generalization.
    \item We validate the efficacy of short-horizon early detection, using the first 5-min of trading data to forecast 1-hour outcomes, specifically within the high-throughput, liquidity-driven microstructure of Solana, showing that reliable predictions can be achieved using only tabular liquidity and trading dynamics without relying on smart contract code analysis.
\end{itemize}

\section{Background: Solana vs Ethereum Memecoins}

Memecoins have emerged as a significant segment within the cryptocurrency landscape, characterized by their blend of humor, speculative appeal, and community engagement \cite{Yousaf2023}. This section explores the contrasting community and technological ecosystems of Solana and Ethereum, two prominent blockchain platforms that support memecoins.

Solana memecoins benefit from the platform's high throughput and generally low transaction costs, attracting traders seeking rapid and cost-effective transactions \cite{Yakovenko2018}. In contrast, Ethereum memecoins leverage the platform's established network and large user base. Ethereum's transition to proof-of-stake improved its consensus mechanism, while its modular scaling architecture can enhance flexibility but introduces additional coordination across layers \cite{Buterin2014}.

Both ecosystems face the risk of rug pulls, in which developers exploit investors and abandon projects or withdraw liquidity \cite{Mazorra2022}. Ethereum's extensive DeFi infrastructure and the ease of launching tokens via the ERC-20 standard create opportunities for such scams \cite{Vogelsteller2015,Angelo2020}. Although Ethereum smart contracts enable decentralized and transparent transactions, access-control mechanisms such as whitelisting can be abused to restrict trading and favor privileged addresses.

\begin{figure}[h]
    \centering
\lstset{language=Solidity}
\begin{lstlisting}[numbers=left, firstnumber=467, linewidth=0.45\textwidth, xleftmargin=12pt, basicstyle=\scriptsize\ttfamily]
function fullWhitelist(address target) public onlyOwner {
    authorizations[target] = true;
    isFeeExempt[target] = true;
    isTxLimitExempt[target] = true;
    isInternal[target] = true;
}
\end{lstlisting}
    \vspace{-0.3cm}
    \caption{Ichimoku Inu’s \texttt{fullWhitelist} function}
    \label{fig:fullWhitelist}
\end{figure}

Most Ethereum memecoins follow the ERC-20 standard, sometimes with modified code. Whitelisting uses Solidity mappings to restrict access to approved addresses and support KYC/AML compliance in token sales and NFT minting \cite{Ermilov2017,Weber2019,Wronka2022}. However, scam tokens may exploit whitelists to favor developer-controlled addresses, allowing them to trade freely while restricting others through blocked transactions or excessive fees \cite{KoinX2024}. Although blockchain provides transparency and smart contract bytecode can often be reverse-engineered \cite{Bitfury2016,Angelo2023,Angelo2024}, many Ethereum memecoins lack publicly available source code, making malicious mechanisms harder to identify and undermining transparency and fairness in DeFi.

\begin{figure}[h]
    \centering
\lstset{language=Solidity}
\begin{lstlisting}[numbers=left, firstnumber=336, linewidth=0.45\textwidth, xleftmargin=12pt, basicstyle=\scriptsize\ttfamily]
function _transferFrom(address sender, address recipient, uint256 amount) internal returns (bool) {
    if (inSwapAndLiquify) {
        return _basicTransfer(sender, recipient, amount);
    }
    if (!authorizations[sender] && !authorizations[recipient]) {
        require(tradingOpen, "");
    }

    require(amount <= _maxTxAmount || isTxLimitExempt[sender], "TX Limit");
    if (isPair[recipient] && !inSwapAndLiquify && swapAndLiquifyEnabled && _balances[address(this)] >= swapThreshold) {
        marketingAndLiquidity();
    }
    if (!launched() && isPair[recipient]) {
        require(_balances[sender] > 0, "");
        launch();
    }

    // Exchange tokens
    _balances[sender] = _balances[sender].sub(amount, "");

    if (!isTxLimitExempt[recipient] && restrictWhales) {
        require(_balances[recipient].add(amount) <= _walletMax, "");
    }

    uint256 finalAmount = !isFeeExempt[sender] && !isFeeExempt[recipient] ? extractFee(sender, recipient, amount) : amount;
    _balances[recipient] = _balances[recipient].add(finalAmount);

    emit Transfer(sender, recipient, finalAmount);
    return true;
}
\end{lstlisting}
    \vspace{-0.3cm}
    \caption{Ichimoku Inu's \texttt{\_transferFrom} function}
    \label{fig:transferFrom}
\end{figure}

The key difference between legitimate Ethereum memecoins such as Pepe Token and whitelist scams such as Ichimoku Inu lies in their smart contract logic. Unlike Pepe Token, Ichimoku Inu uses dynamic whitelisting (Figures~\ref{fig:fullWhitelist} and~\ref{fig:transferFrom}) to favor privileged addresses. Non-whitelisted users may be blocked from trading when \texttt{tradingOpen} is false and remain subject to transaction and whale restrictions, while whitelisted addresses are exempt. In contrast, Solana Program Library (SPL) standard supports fungible and non-fungible tokens through a modular framework, making whitelist abuse less feasible \cite{Solana2025}.

While both Solana and Ethereum offer fertile grounds for memecoin development, their distinct technological frameworks and community dynamics shape their respective ecosystems.

\section{Related Work}

The application of machine learning to the detection of financial fraud has been extensively surveyed~\cite{yang2026review}, covering tree-based models, deep learning, and graph methods in traditional domains such as credit card fraud, loan fraud, and anti-money laundering. However, rug pulls in decentralized memecoin markets present unique challenges: extreme class imbalance, rapid token lifecycles, and chain-specific microstructures not fully addressed by traditional fraud detection frameworks.

The cryptocurrency landscape has witnessed significant developments over recent years, characterized by both innovation and deception. Fraudulent activities such as Ponzi schemes~\cite{Chen2018b,Chen2019a,Galletta2023}, pump-and-dump tactics~\cite{Chadalapaka2022,Bello2023}, and rug pulls~\cite{Mazorra2022,Srifa2025,yaremus2025ton} have become increasingly prevalent. The decentralized nature of the market facilitates these scams, with rug pulls particularly leaving investors with worthless tokens after developers withdraw liquidity. The anonymity and lack of regulation~\cite{Feinstein2021,Chaleenutthawut2024} further complicate detection and prevention.

During the initial coin offering (ICO) era~\cite{Fenu2018}, a surge in fraudulent schemes emerged, reminiscent of the current memecoin trend. ICOs analysis has highlighted vulnerabilities that persist in the cryptocurrency sphere~\cite{Howell2020}. These insights are especially relevant as the modern memecoin frenzy, particularly on platforms such as Uniswap V2~\cite{Adams2018,Adams2020}, has increased rug pull risk. Memecoins~\cite{Stencel2023,CMCResearch2024}, with their viral marketing and speculative appeal, are prime targets for fraudulent activities.

Recent advances in cryptocurrency fraud detection have focused on identifying rug pulls, especially on the Ethereum blockchain~\cite{Buterin2014}. Mazorra et al.~\cite{Mazorra2022} developed a comprehensive labeled dataset of 26,957 tokens identified as scams on Uniswap, providing a theoretical classification of rug pull types--simple, sell, and trap-door--and introducing tools to identify them. Their work emphasizes token distribution metrics, such as the Herfindahl--Hirschman Index, uncovering that 90\% of tokens using lock contracts like Unicrypt eventually become malicious. Srifa et al.~\cite{Srifa2025} extended this to Uniswap V3, analyzing 7,450 tokens (3,212 normal, 581 rug pulls) and shifting focus from detection to timing prediction using time-series features like token volume and transaction frequency.

Beyond liquidity-based signals, behavioral pattern profiling has been explored for early warning. Cao et al.~\cite{cao2026early} constructed 12 token-level features based on three wash-trading patterns for BSC meme tokens, achieving high PR-AUC with Random Forest. Their work emphasizes lead-time analysis (mean 3.8 hours of early warning) and interpretable feature ablation. Although their focus is BSC and wash-trade signals, our approach targets Solana and liquidity-based labels (TVL/Idle), suggesting that combining behavioral and liquidity signals could improve cross-chain detection.

Beyond feature-based machine learning, graph neural networks have shown promise in modeling token interaction patterns. Wu et al.~\cite{wu2025rugscreener} proposed RugScreener, a temporal GNN architecture that captures dynamic transaction graphs for ERC-20 tokens on Ethereum. While their approach leverages relational structure, our work focuses on interpretable, tabular features derived from liquidity dynamics and trading behavior--offering a complementary, lower-complexity alternative suitable for real-time deployment on high-throughput chains like Solana. 

Beyond Ethereum, recent work has extended rug pull detection to other blockchains. Yaremus et al.~\cite{yaremus2025ton} developed a machine learning framework for The Open Network (TON), analyzing data from STON.fi and DeDust DEXs. Like our work, they compare two rug pull definitions--TVL-based liquidity withdrawal and idle-based trading cessation--and demonstrate that Gradient Boosting models can identify scams within five minutes of trading. Their key finding that feature distributions differ significantly across DEXs directly motivates our cross-platform analysis of Pumpfun and Raydium.

While most prior rug pull detection research targets Ethereum~\cite{Mazorra2022,Srifa2025}, Hu et al.~\cite{hu2026memetrans} recently released MemeTrans, a dataset of 40k Solana memecoins with features spanning trading activity, holding concentration, and bundle-level entity resolution. Their annotation combines statistical indicators with the detection of manipulation-patterns, achieving a 56.1\% reduction in simulated financial loss. Our work complements this by scaling to 160$\times$ more tokens (6.4M vs. 40k), using liquidity-based rug pull definitions (TVL drop / prolonged idleness) rather than launchpad-phase heuristics, and evaluating cross-platform generalization between Pumpfun and Raydium.

\section{Methodology}
\subsection{Data Collection and Preprocessing}
The data used for this research was collected directly from blockchain nodes and included transaction records, especially focusing on the minting of new tokens. Data from two decentralized exchanges (DEXes) were analyzed: Raydium and PumpFun. It is important to note that these platforms are interconnected: tokens that achieve a market capitalization of \$69,000 on PumpFun are automatically migrated to Raydium for further trading. And from November 2024 to  2025, the crypto market witnessed Memecoin's second explosive growth. The key event in this wave was the deep integration of AI concepts with meme culture.

\noindent For each DEX, the following data columns were extracted and standardized:

\begin{table}[ht]
\centering
\scriptsize
\caption{Data columns from Raydium and PumpFun.}
\begin{tabular}{lll}
\toprule
\multicolumn{2}{c}{\textbf{PumpFun}} & \textbf{Raydium} \\
\midrule
block\_time & slot & slot \\
tx\_idx & creator & tx\_idx \\
name & symbol & address \\
url & mint & signature \\
bundle\_size & gas\_used & block\_time \\
amount\_of\_instructions & amount\_of\_lookup\_reads & signing\_wallet \\
amount\_of\_lookup\_writes & bundle\_structure & direction \\
bundled\_buys & bundled\_buys\_count & base\_coin \\
dev\_balance & creation\_ix\_index & quote\_coin \\
curve\_address & pf\_program\_index & base\_coin\_amount \\
direct\_pf\_invocation & version & quote\_coin\_amount \\
mayhem\_mode & token\_program & base\_pool\_before \\
signature & parent\_program & quote\_pool\_before \\
signing\_wallet & direction & base\_pool\_after \\
base\_coin & base\_coin\_amount & quote\_pool\_after \\
quote\_coin\_amount & virtual\_token\_balance\_after & serum\_market\_id \\
virtual\_sol\_balance\_after & provided\_gas\_fee & raydium\_market\_id \\
provided\_gas\_limit & fee & provided\_gas\_fee \\
consumed\_gas & & provided\_gas\_limit \\
& & fee \\
& & consumed\_gas \\
& & parent\_program \\
\bottomrule
\end{tabular}
\label{tab:dataset_columns}
\end{table}

The data was collected over a 7-month period for each DEX (Nov. 30, 2024 -- Jun. 30, 2025). Given the high transaction activity on the Solana blockchain, this timeframe provided sufficient observations for robust modeling. We performed preprocessing to construct a dataset of memecoin-SOL trading pairs. Table \ref{tab:dex_summary} demonstrates the scale of our token dataset and the wealth of transaction records.

\begin{table}[ht]
\centering
\scriptsize
\caption{Dataset statistics for Raydium and PumpFun.}
\begin{tabular}{lcc}
\toprule
\textbf{DEX} & \textbf{Number of tokens} & \textbf{Total Swap Transactions} \\
\midrule
Raydium & 97965 & $5.923951\times 10^{8}$ \\ 
PumpFun & 6304235 & $4.226951\times 10^{8}$ \\
\bottomrule
\end{tabular}
\label{tab:dex_summary}
\end{table}

To effectively model token behavior, additional features were derived from the raw data, capturing both token performance on the DEXs and token creator-specific attributes. These features were aggregated into a structured dataset to serve as input to the machine learning model. Table \ref{tab:features} lists the basic data features required for this study.

\begin{table}[ht]
\centering
\scriptsize
\setlength{\tabcolsep}{2pt}
\caption{Calculated features to observe token activity.}
\begin{tabular}{@{}>{\raggedright\arraybackslash}p{0.47\columnwidth}@{\hspace{2pt}}>{\raggedright\arraybackslash}p{0.49\columnwidth}@{}}
\toprule
\textbf{Feature} & \textbf{Description} \\ 
\midrule
count\_tx & Total number of transactions associated with the token. \\ 
purchase\_percentage & Percentage of transactions classified as token purchases. \\ 
sale\_percentage & Percentage of transactions classified as token sales. \\ 
unique\_buyers & Number of unique wallets that purchased the token. \\ 
unique\_sellers & Number of unique wallets that sold the token. \\ 
total\_sol\_value & Total value of transactions in SOL associated with the token. \\ 
sell\_sol\_value & Total SOL value from token sale transactions. \\ 
buy\_sol\_value & Total SOL value from token purchase transactions. \\ 
buy\_sell\_cnt\_ratio & Ratio of buy transaction count to sell transaction count. \\ 
buy\_sell\_value\_ratio & Ratio of buy SOL value to sell SOL value. \\ 
\shortstack[l]{price\_change\_first\\to\_3\_blocks} & Price change of the token during its first three blocks of activity. \\
\shortstack[l]{price\_change\_first\\to\_last} & Price change from first to last trade. \\
buy\_price\_std & Standard deviation of buy prices. \\ 
sell\_price\_std & Standard deviation of sell prices. \\ 
first\_buy\_time & Timestamp of the first buy transaction. \\ 
first\_sell\_time & Timestamp of the first sell transaction. \\ 
min\_pool\_info\_time & Timestamp of the earliest pool information. \\ 
max\_pool\_info\_time & Timestamp of the latest pool information. \\ 
last\_trade\_time & Timestamp of the last trade. \\ 
max\_price & Maximum price. \\ 
min\_price & Minimum price. \\ 
min\_price\_after\_max & Minimum price after the maximum price. \\ 
first\_price & Price at the first trade after mint. \\ 
\bottomrule
\end{tabular}
\label{tab:features}
\end{table}

% 2
\subsection{Rug Pull Target Variable Definition}

Following prior work on multi-DEX rug pull detection~\cite{yaremus2025ton}, this study adopts both Idle and TVL approaches \cite{detecting-rug-pulls} to define rug pulls, each corresponding to a distinct failure mode in decentralized markets.

\textbf{TVL Approach:} The Total Value Locked is an important investment indicator. It is typically the total value of the LPs over a certain period of time. The higher the total value, the stronger its liquidity.
\begin{equation*}
\mathrm{TVL}_t = \sum_{p \in \mathcal{P}} \left( P^{(base)}_{p,t} \cdot Q^{(base)}_{p,t} + P^{(quote)}_{p,t} \cdot Q^{(quote)}_{p,t} \right)
\end{equation*}

TVL rug pulls are one of the most common scams in meme investing. They typically use false advertising, promises of high returns, and marketing pump-and-dump schemes to attract users to deposit assets. At this point, TVL can experience a rapid increase in value in a short period, followed by an immediate withdrawal of funds from the liquidity pool, causing TVL to drop by 99\%.

\textbf{Idle Approach:} The idle approach defines a rug pull in which no transactions occur for an extended period. \\
\begin{equation*}
\mathrm{Idle}_t = \mathbb{I}\left( \sum_{\tau = t-w}^{t} V_\tau = 0 \right)
\end{equation*}

When a token enters a dormant state for a long time, it effectively loses liquidity. Specifically, when the inactivity time ratio is high, the market exhibits low trading intensity, and thus low liquidity.

\textbf{TVL OR Idle Rug Pull:} Based on the above definition, when a token triggers a TVL Rug Pull condition or becomes Idle during trading, the token is classified as a high-risk token. The specific determination rules are as follows:

The maximum drawdown depth (MDD) of TVL is defined as follows:
$
\mathrm{MDD}_t = \frac{\mathrm{TVL}_t - \max_{\tau \le t} \mathrm{TVL}_\tau}{\max_{\tau \le t} \mathrm{TVL}_\tau}.
$

This value is between [-1, 0]. A negative value indicates that TVL has fallen from its peak, and the larger the absolute value, the more severe the liquidity withdrawal. Therefore, the Rug Pull event indicator variable is defined as follows: \\
$
\mathrm{Rug}_t = \mathbb{I}\left( (\mathrm{MDD}_t < -\theta) \lor (\mathrm{Idle}_t > \Delta t) \right).
$

The thresholds $\theta$ and $\Delta t$ are empirically selected based on the distributional analysis of TVL drops and durations of inactivity.

\subsection{Rolling Time Series Cross-Validation}
The construction of training and test sets directly impacts the reliability of model evaluation. Since Rug Pull detection is a time-dependent task, random partitioning or standard K-fold cross-validation introduces look-ahead bias, exposing the model to future market information during training and leading to overly optimistic evaluation results. Therefore, this study employs forward rolling window cross-validation based on token issuance time, reserving the last time window as an independent test set. Another reason for using time partitioning is that Rug Pull fraud exhibits concept drift characteristics, with its on-chain features constantly changing with the market. Compared to standard K-fold cross-validation, which mixes data from different time periods, a fixed-length rolling window utilizes only the most recent 3 months of data for training.

Furthermore, because Rug Pulls are a minority class event, this study implements a positive sample guarantee mechanism. When the number of positive samples in the validation set is insufficient, the validation window is automatically expanded to ensure that each validation fold contains at least 5 Rug Pull samples, thereby guaranteeing the statistical significance of the evaluation results.

This partitioning method follows the process of "historical training, future verification" to avoid data leakage and ensure that the final test set is used only for one-time evaluation, thereby improving the credibility of the experimental results.

% 3
\subsection{Detection Model and Parameter Optimization}

The primary objective of this study is to develop a detection learning model capable of predicting token behavior. We simplified the complex task of rug pull detection into a binary classification problem, predicting a token's future behavior using only the first 5 minutes of trading data. For traditional machine learning, we selected Random Forest, XGBoost, and MLP. We utilized Optuna (Bayesian search) to optimize hyperparameters across all folds, subsequently retraining the models with the average optimal parameters. For transformer models, we choose FT-Transformer, TabTransformer, AutoInt. They represent three different ideas for tabular data modeling. We incorporated a cross-entropy loss function into our experiments to control the training and understanding of the model.

Then, we evaluated the model's predictive performance on different platforms (Pumpfun and Raydium) under transfer training and fusion training to identify potential anisotropies in model performance. 

% The following conclusions were drawn: TO DO

% 4
\subsection{Experimental Prediction Framework}

Rug Pull is defined as a fraudulent event in which the developers of a project abandon it, leading to a rapid and substantial decline in the token's value. In this study, the target labels are binary indicators of whether a TVL rug pull or an Idle rug pull has occurred at 1 hour.

\begin{figure}[ht]
    \centering
    \includegraphics[width=\linewidth]{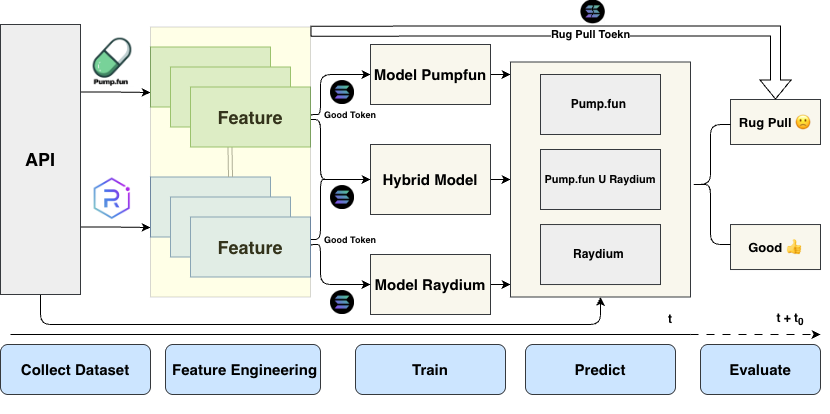}
    \caption{The Memecoin Future Prediction Framework uses Time $t$ to predict $t+t_{0}$.}
    \label{fig:network}
\end{figure}

This paper constructs a basic experimental prediction framework Figure \ref{fig:network} for DeFi scenarios. The framework first collects unified transaction data related to PumpFun and Raydium through blockchain indexer, and performs feature engineering at the data layer to extract multi-dimensional features such as price dynamics, liquidity changes, and trading behavior. At the model layer, dedicated models for PumpFun and Raydium are designed, respectively, and a hybrid model integrating the distribution characteristics of the two types of data is further constructed to improve cross-platform generalization capabilities. In the prediction phase, new input samples are classified and discriminated, dividing tokens into potential rug pull Tokens or Good Tokens. Finally, in the evaluation module, the model performance is systematically verified using multiple indicators. The preset window of the experiment is t = 5 min, predicting the likelihood of a rug pull within a 1-hour horizon. Specifically, the input features are strictly selected from within the 5-min window to prevent data leakage into the 60-min prediction timeframe.

% 5
\subsection{Model Evaluation Metrics}

In the crypto market, over 80\% of Memecoins experience rug pulls. This research is based on unbalanced data from the real world of the cryptocurrency market. For unbalanced label recognition tasks, the accuracy (ACC) does not accurately reflect the performance of the classifier. When the data distribution is unbalanced, it often causes the output of the classifier to tend to the Rug class, which will have a higher classification accuracy, but performs poorly in the minority class Non-Rug.

To solve this problem, the study adopts evaluation indicators specifically designed for imbalanced classification. Rather than reporting all standard metrics, we focus on the three most robust indicators for detecting the minority class (Rug Pulls): the F1-score for the positive class, the Matthews Correlation Coefficient (MCC), and the Area Under the Precision-Recall Curve (AUCPRC).

\begin{itemize}
    \item \(F1\text{-score (Positive Class)} = 2 \cdot \frac{\text{Recall} \times \text{Precision}}{\text{Recall} + \text{Precision}}\), where \(\text{Recall} = \frac{TP}{TP+FN}\) and \(\text{Precision} = \frac{TP}{TP+FP}\)
    \item \(MCC = \frac{TP \times TN - FP \times FN}{\sqrt{(TP+FP)(TP+FN)(TN+FP)(TN+FN)}}\)
    \item AUCPRC = Area under the Precision-Recall (PR) curve
\end{itemize}

These three metrics together provide a comprehensive and unbiased evaluation framework for the severe class imbalance present in rug pull detection.

\section{Numerical Experiments}

\subsection{Data Analysis}
 
 The data contains addresses for 6.4 million Memecoins. For all Memecoins, the study collected transaction data from the first hour for preprocessing, and Table \ref{tab:dex_detailed_summary} illustrates some basic data characteristics.

\begin{table}[ht]
\centering
\scriptsize
\caption{Descriptive statistics comparison between Pumpfun and Raydium.}
\begin{tabular}{lcc}
\toprule
\textbf{Metric} & \textbf{Pumpfun (5mins)} & \textbf{Raydium (5mins)} \\
\midrule
Total Transactions & $2.47 \times 10^{8}$ & $3.42 \times 10^{7}$ \\ 
Avg Tx & $3.93 \times 10^{1}$ & $5.88 \times 10^{2}$ \\ 
Median Tx & $1.20 \times 10^{1}$ & $3.85 \times 10^{2}$ \\ 
Total Volume (SOL) & $1.28 \times 10^{17}$ & $3.08 \times 10^{16}$ \\ 
Avg Volume & $2.04 \times 10^{10}$ & $5.29 \times 10^{11}$ \\ 
Avg Buyers & $1.83 \times 10^{1}$ & $1.54 \times 10^{2}$ \\ 
Avg Sellers & $1.30 \times 10^{1}$ & $8.42 \times 10^{1}$ \\ 
Avg Buy Ratio & $6.21 \times 10^{-1}$ & $7.17 \times 10^{-1}$ \\ 
Avg Sell Ratio & $3.94 \times 10^{-1}$ & $3.00 \times 10^{-1}$ \\ 
Avg Price Change & $1.56 \times 10^{2}$ & $5.08 \times 10^{5}$ \\ 
Avg Buy Std & $2.62 \times 10^{-2}$ & $1.99 \times 10^{5}$ \\ 
Avg Sell Std & $4.95 \times 10^{-6}$ & $6.25 \times 10^{2}$ \\ 
Avg Last Trade time & $1.62 \times 10^{2}$ & $2.97 \times 10^{2}$ \\  
\bottomrule
\end{tabular}
\label{tab:dex_detailed_summary}
\end{table}

From these characteristics, we can see that Pumpfun's trading model is "multiple pools and small trades," while Raydium's trading model is "few pools and large trades."

\subsection{Rug Pull Labels Analysis}

Prolonged idleness indicates that a token is not being continuously or effectively traded, while a sharp TVL drop signifies that a token lacks reliable liquidity. This study extracts rug pull labels from the feature data and performs statistical analysis to facilitate a better understanding of the machine learning models' behavior. We define a rug pull label as a token whose TVL falls below\textbf{ 99\%} or whose idle time exceeds \textbf{80\%} of its lifetime without trading.

\begin{figure}[ht]
    \centering
    
    % Left Subfigure (Top-Aligned)
    \begin{subfigure}[t]{0.49\linewidth}
        \centering
        \includegraphics[width=\linewidth]{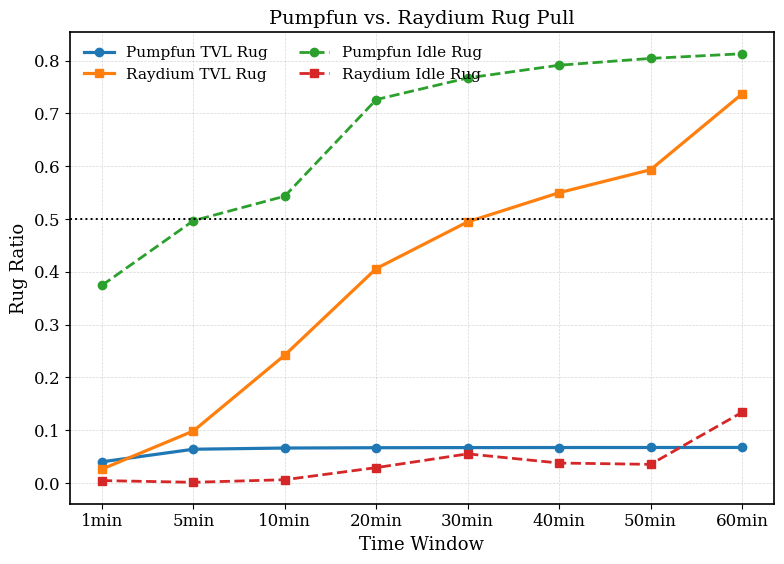}
        \caption{TVL and Idle label}
        \label{fig:labels1}
    \end{subfigure}
    \hfill % Pushes the next subfigure to the right
    % Right Subfigure (Top-Aligned)
    \begin{subfigure}[t]{0.49\linewidth}
        \centering
        \includegraphics[width=\linewidth]{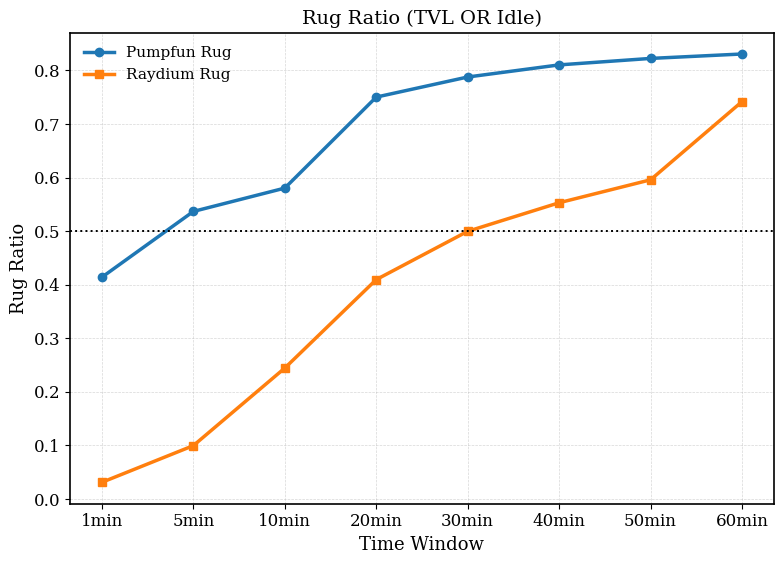}
        \caption{Overall Rug Pull label}
        \label{fig:labels2}
    \end{subfigure}
    
    \caption{Distribution of rug pull labels on Pumpfun and Raydium. (a) Rug pull ratio over time under TVL/idle labels on Pumpfun and Raydium. (b) Rug pull ratio over time under TVL OR idle labels on Pumpfun and Raydium.}
    \label{fig:labels_combined}
\end{figure}

The results of this research are analyzed using these rug pull labels across different DEXs. Figure \ref{fig:labels1} and Figure \ref{fig:labels2} show that these results indicate that signals based on trading idle time are particularly effective in identifying highly speculative markets such as Pumpfun, while signals based on TVL more consistently indicate liquidity withdrawals in mature markets such as Raydium, although this indication is somewhat delayed. Overall, combining these two criteria improves robustness and enables earlier detection of such behavior across different DEXs. When acquiring 5-minute data features, we drop data that has already been rug-pulled within that 5-min period. This reduces computational load and better reflects market trading patterns. Table \ref{tab:testset} shows the distribution of the processed test dataset.

\begin{table}[ht]
\centering
\scriptsize
\caption{Label distribution of the testset between Pumpfun and Raydium.}
\begin{tabular}{lcc}
\toprule
\textbf{Defi} & \textbf{Rug Pull} & \textbf{Non Rug Pull} \\
\midrule
Raydium & $43835$ & $9711$ \\ 
Pumpfun & $2931946$ & $1893539$ \\
\bottomrule
\end{tabular}
\label{tab:testset}
\end{table}

\subsection{Detection Model Analysis}

Statistical analysis of the labeled data reveals a significant upward trend in the rug pull risk of the Memecoin project over time. Specifically, the risk is relatively low in the early stages of trading, but the probability of liquidity withdrawal or price collapse gradually increases over time. Therefore, this study uses a 5-minute prediction window and a 1-hour alert window to evaluate the performance of different machine learning models in the rug pull prediction task.

Combining the experimental results in Table \ref{tab:performance_groups}, it can be further observed that different data domains (such as Raydium, PumpFun, and their cross-DEX transfer) have a significant impact on model performance. In general, training within the same domain (such as Raydium or PumpFun) shows stable performance, with MCC scores around 0.25--0.36 across most models. However, cross-DEX prediction (such as Raydium\(\to\)PumpFun) leads to a significant performance drop, with MCC values approaching zero or becoming negative across nearly all models, indicating substantial distributional differences between the two DEX platforms. Among the models, tree-based approaches (Random Forest and XGBoost) consistently outperform neural network-based models (MLP, FTTransformer, TabTransformer, and AutoInt) in cross-domain scenarios, demonstrating stronger robustness to domain shifts.

Interestingly, when models are trained on the fused dataset (Raydium U PumpFun) and evaluated on either domain, they achieve results comparable to or even better than single-domain training on most metrics. For instance, in the Fusion\(\to\)PumpFun direction, XGBoost and Random Forest achieve MCC values of 0.3947 and 0.3942, respectively, which are superior to the PumpFun\(\to\)PumpFun domain-specific results (0.3562 and 0.3558). This suggests that multi-source data integration effectively improves the model's generalization ability across DEX environments. In particular, XGBoost and Random Forest exhibit the most robust performance on AUCPRC and F1 metrics when leveraging fused data, validating the effectiveness of the data fusion strategy in the rug pull prediction task. However, it should be noted that Transformer-based models (FTTransformer, TabTransformer, AutoInt) show limited improvement from fusion on the Raydium testset (MCC values of -0.0444, -0.0543, and 0.0247, respectively), indicating that advanced neural architectures do not necessarily guarantee better cross-domain transferability in this specific financial prediction context. Overall, these findings highlight the importance of domain-aware model selection and multi-source data integration for building reliable DeFi risk prediction systems. We hypothesize that Raydium's more mature, lower-noise trading data provides a stronger foundational signal of legitimate liquidity, which helps the model generalize better to the noisier PumpFun environment when fused.

\begin{table}[htbp]
\centering
\scriptsize
\caption{Rug Pull Prediction Results (TVL OR Idle). TVL OR Idle denotes rug pulls identified by either 99\% TVL drop OR 80\% idle time. Best results per column are \textbf{bolded}.}
\label{tab:performance_groups}

\renewcommand{\arraystretch}{1.1}
\setlength{\tabcolsep}{3.5pt}
\begin{tabular}{lccccccc}
\toprule
\multirow{2}{*}{Method} &
\multicolumn{7}{c}{TVL OR Idle} \\
\cmidrule(r){2-8}
 & R & R$\rightarrow$P & P & P$\rightarrow$R & R $\cup$ P & F$\rightarrow$P & F$\rightarrow$R\\
\midrule

\multicolumn{8}{l}{\textbf{Random Forest}}\\
F1(1)     & 0.7524 & 0.7207 & 0.7785 & 0.6106 & \textbf{0.7838} & 0.7877 & \textbf{0.7408} \\  
MCC       & \textbf{0.2874} & -0.0434 & 0.3558 & -0.2433 & \textbf{0.3938} & 0.3942 & \textbf{0.2246} \\  
AUCPRC    & \textbf{0.7578} & 0.5303 & \textbf{0.7634} & 0.5435 & 0.7897 & 0.7978 & \textbf{0.6040} \\ 
\midrule

\multicolumn{8}{l}{\textbf{XGBoost}}\\
F1(1)     & \textbf{0.7532} & 0.7238 & \textbf{0.7810} & 0.3778 & 0.7833 & \textbf{0.7885} & 0.7288 \\  
MCC       & 0.2914 & -0.0026 & 0.3562 & -0.0664 & 0.3888 & \textbf{0.3947} & 0.1543 \\ 
AUCPRC    & 0.7548 & 0.5771 & 0.7564 & \textbf{0.5792} & \textbf{0.8003} & \textbf{0.8011} & \textbf{0.6426} \\  
\midrule

\multicolumn{8}{l}{\textbf{MLP}}\\
F1(1)     & 0.7482 & 0.7221 & 0.7800 & 0.7068 & 0.7333 & 0.7362 & 0.6916 \\ 
MCC       & 0.2663 & -0.0427 & 0.3517 & \textbf{0.0865} & 0.3143 & 0.3144 & -0.1121 \\ 
AUCPRC    & 0.7379 & 0.5019 & 0.7457 & 0.5317 & 0.7437 & 0.7588 & 0.5833 \\ 
\midrule

\multicolumn{8}{l}{\textbf{FTTransformer}}\\
F1(1)     & 0.7510 & 0.7247 & 0.7796 & \textbf{0.7130} & 0.7674 & 0.7715 & 0.7174 \\  
MCC       & 0.2807 & -0.0508 & \textbf{0.3649} & 0.0387 & 0.3232 & 0.3257 & -0.0444 \\  
AUCPRC    & 0.7506 & 0.4654 & 0.7428 & 0.5576 & 0.7297 & 0.7470 & 0.5474 \\ 
\midrule

\multicolumn{8}{l}{\textbf{TabTransformer}}\\
F1(1)     & 0.6286 & \textbf{0.7408} & 0.7621 & 0.3530 & 0.7480 & 0.7543 & 0.6295 \\  
MCC       & 0.2508 & \textbf{0.0991} & 0.2867 & -0.1479 & 0.2580 & 0.2719 & -0.0543 \\  
AUCPRC    & 0.6997 & \textbf{0.6628} & 0.7168 & 0.5224 & 0.7003 & 0.7093 & 0.5686 \\ 
\midrule

\multicolumn{8}{l}{\textbf{AutoInt}}\\
F1(1)     & 0.7501 & 0.7294 & 0.7797 & 0.6992 & 0.7694 & 0.7692 & 0.6851 \\  
MCC       & 0.2758 & -0.0575 & 0.3491 & -0.0946 & 0.3339 & 0.3307 & 0.0247 \\  
AUCPRC    & 0.7346 & 0.4775 & 0.7492 & 0.5137 & 0.7249 & 0.7354 & 0.5854 \\ 

\bottomrule
\end{tabular}
\vspace{0.5em}

\scriptsize \textit{Note: R = Raydium (train/test on Raydium), P = PumpFun (train/test on PumpFun), \\ F = Fusion (train on R $\cup$ P), $\rightarrow$ = transfer learning direction, $\cup$ = combined dataset.}
\vspace{-0.3cm}
\end{table}

\section{Limitations And Discussion}

This study should be considered as a baseline exploration of rug pull prediction in decentralized exchanges. Although several models achieve relatively strong performance (e.g., AUCPRC exceeding 0.8 in certain settings), these results are not yet sufficient for real-world deployment, especially in high-risk financial environments where false negatives carry substantial investor losses.

Our results reveal substantial performance degradation under cross-DeFi settings (Table~\ref{tab:performance_groups}), indicating distribution shift between platforms. This aligns with the findings of Yaremus et al. on TON~\cite{yaremus2025ton}, where cross-DEX transfer learning also suffered from feature distribution mismatch. The pattern suggests that platform-aware adaptation--rather than naive data fusion--may be necessary for robust cross-DEX rug pull detection.

Our dataset (6.4M tokens) allows a more robust statistical analysis of rare rug pull patterns. However, MemeTrans provides richer bundle-level features for entity resolution, suggesting a promising direction for future feature engineering. Integrating bundle detection could help identify coordinated manipulation that our current feature set may miss.

The current study relies on liquidity dynamics and price volatility features (Table~\ref{tab:features}), reflecting Solana's distinct market microstructure. Cao et al.~\cite{cao2026early} demonstrate that trade-level wash-trading features are primary drivers of detection performance in BSC. Future work could fuse both signal types--liquidity-based and behavioral--for more robust cross-chain generalization.

We employ standard tabular models, which may not fully capture the temporal dynamics, liquidity evolution, and microstructure patterns inherent in DeFi markets. More advanced modeling techniques---such as temporal GNNs~\cite{wu2025rugscreener} or sequence-based transformer architectures---could better represent these complex dynamics, albeit at a higher computational cost.

\section{Conclusion And Future Work}

This paper presents the first large-scale systematic study of rug pull prediction on Solana memecoins. We implement and compare two rug pull definitions--TVL-based liquidity withdrawal and Idle-based trading cessation--within a unified study, following the methodology validated on TON~\cite{yaremus2025ton}. We evaluate the generalization of the model between Pump.fun and Raydium, revealing a significant distribution shift that challenges naive data fusion approaches. We demonstrate that Gradient Boosting models can identify rug pulls within 5 minutes of trading, with XGBoost achieving the highest cross-platform generalization on fused data.

The experimental results highlight that different platforms exhibit distinct trading characteristics: PumpFun is characterized by high-frequency, small-volume trades, whereas Raydium tends to involve lower-frequency, large-volume trades. These structural differences lead to varying data distributions and directly affect model performance.

Data fusion proved feasible and promising, as the combination of data from multiple platforms improved model robustness and partially mitigated domain-specific biases. This suggests the potential for developing more generalizable cross-platform prediction models.

For future work, several directions are worth exploring. First, advanced architectures such as temporal GNNs~\cite{wu2025rugscreener} could be introduced to capture dynamic transaction patterns. Second, feature enrichment could integrate bundle-level entity resolution~\cite{hu2026memetrans} and wash-trading patterns~\cite{cao2026early} with our liquidity-based features. Third, shorter prediction horizons below 5 minutes could be investigated for earlier warning, balancing accuracy against actionable lead time.

As Solana continues to dominate the memecoin market~\cite{CMCResearch2024}, robust fraud detection mechanisms become increasingly vital. This work provides a foundation for protecting investors and maintaining trust in the Solana DeFi ecosystem.

\bibliographystyle{IEEEtran}
\bibliography{referencesLocal}

\end{document}